# rl_rob: A transparent, from-scratch Python library for teaching reinforcement learning from tabular methods to deep policy gradients

Abdulrahman Altahhan
*School of Computer Science, University of Leeds, Leeds LS2 9JT, United Kingdom*
Email: A.Altahhan@leeds.ac.uk

**Abstract**

rl_rob is an open-source Python library for teaching reinforcement learning. It implements the arc of a modern course, from tabular temporal-difference learning through linear function approximation to deep value-based and policy-gradient methods, with every algorithm written from first principles so that its update rule is visible in a few lines of code. All agents share one interface, so learners can compare methods directly across stages and trace a deep reinforcement learning update from the temporal-difference error to the weight change. The library supports teaching and assessment in which students implement, modify and explain update rules, rather than only produce working code.

**Code metadata**

| Nr. | Code metadata description | Metadata |
|---|---|---|
| C1 | Current code version | v1.0.0 |
| C2 | Permanent link to code/repository used for this code version | https://github.com/AltahhanAi/rl_rob |
| C3 | Permanent link to Reproducible Capsule | Not applicable |
| C4 | Legal Code License | GNU Affero General Public License v3.0 (AGPL-3.0) |
| C5 | Code versioning system used | git |
| C6 | Software code languages, tools, and services used | Python; NumPy, Matplotlib, PyTorch, Gymnasium, Jupyter; GitHub Actions; Google Colab |
| C7 | Compilation requirements, operating environments and dependencies | Python 3.9 or later on Linux, macOS or Windows, or Google Colab; numpy, matplotlib, ipython, tqdm, gymnasium, minigrid, torch, opencv-python-headless, scikit-learn; optional ROS 2 for the Gazebo robot environment |
| C8 | If available, link to developer documentation/manual | https://github.com/AltahhanAi/rl_rob#readme |
| C9 | Support email for questions | A.Altahhan@leeds.ac.uk |

## 1. Motivation and significance

Reinforcement learning (RL) is now a core part of postgraduate curricula in artificial intelligence, yet it remains one of the hardest subjects to teach well. Its central ideas, bootstrapping, credit assignment over time and the interaction between exploration and learning, are easy to state and hard to see. Students meet them first as pseudocode in a textbook [1] and then, if they continue, as calls to production libraries such as Stable-Baselines3 [2] or CleanRL [3]. Between the two there is a gap.

Production libraries are excellent for applying algorithms, but they hide the update rules behind optimisers, automatic differentiation and engineering layers that are not the subject of an introductory course. Educational resources such as Spinning Up [4] concentrate on deep RL and assume that the tabular and linear foundations are already in place.

In practice an instructor faces an uncomfortable choice. Students can be asked to implement every algorithm and its environment from scratch, which consumes much of a course in plumbing, or they can be given a framework in which the learning itself is invisible. Neither serves the aim of a course, which is for students to understand why an algorithm behaves as it does and when one method should be preferred to another.

rl_rob was written to close this gap. It covers the tabular, linear and neural stages of the standard progression [1] within one consistent interface, so that moving from Q-learning to a deep Q-network [5] changes one class name rather than the whole program. In the neural stage, PyTorch [6] supplies only the network layers and their gradients; the temporal-difference target, the replay buffer, eligibility traces and the policy-gradient updates are written out explicitly. A student can therefore trace a deep RL update from the temporal-difference error to the change in the weights, using the same vocabulary and code structure they met in the tabular stage.

The design also responds to a newer pressure on teaching. Where generative AI tools can produce working code for a standard algorithm on request, assessment that rewards working code alone no longer evidences understanding. Because rl_rob supplies the environments, experiment loop and visualisation, instructors can set tasks in which students implement or modify the core update of an algorithm, predict and explain its behaviour, and compare it with alternatives on the same problem. These tasks remain meaningful when code generation is freely available, because the object of assessment is the student's reasoning about the update rule.

## 2. Software description

### *2.1. Software architecture*

rl_rob is a pure Python package of about seventy classes, organised in two layers: algorithms (package rl) and environments (package env). Its dependencies are standard scientific Python libraries; the tabular and linear stages use only NumPy [7] and Matplotlib [8].

The algorithm layer is built on a small hierarchy of base classes in rl.base. A prediction base (MRP) holds what every agent shares: hyperparameters, seeding, the episode loop, logging, and the plotting and animation harness. A control base (MDP) adds action selection and exploration, and a policy-gradient base (PG) adds parameterised policies. Control and policy-gradient behaviour are attached to each stage through class factories, so that the tabular, linear and neural stages reuse the same control logic and differ only in how values or policies are represented and updated. Each concrete algorithm is then a short class whose main content is its update rule. This is the property the library is built around: the code a student reads for Sarsa, for linear Sarsa(λ) or for a neural actor-critic is the update itself, and the surrounding machinery is identical across them.

The modules follow the stages of a course: rl.tabular and rl.dp for tabular and dynamic-programming methods; rl.linear for linear function approximation; rl.neural and rl.nn for neural function approximation, where rl.nn builds configurable convolutional trunks and fully connected heads; and rl.select, which manages repeated runs and comparisons between algorithms.

The environment layer provides gridworlds, mazes, random walks, windy and cliff-walking grids with tabular, linear and neural state encodings (env.grid); wrappers for Gymnasium control tasks [9] with additive discretisation and hashed multi-tiling tile coding (env.gym); and an optional ROS 2 interface for robot navigation in the Gazebo simulator (env.gazebo), which lets the same agents control a simulated mobile robot.

### *2.2. Software functionalities*

The library implements the following methods, each reachable through the same constructor-and-interact pattern.

- Tabular methods: Monte Carlo and TD(0) prediction, n-step methods, Sarsa, Q-learning, expected Sarsa, double Q-learning, Dyna-Q and prioritised sweeping, together with dynamic programming.
- Linear function approximation: prediction and control with features, n-step Sarsa, eligibility traces including true online TD(λ) [10], REINFORCE and actor-critic with softmax and Gaussian policies, applied to Mountain Car, Acrobot, Pendulum and CartPole through discretisation and tile coding.
- Neural function approximation: neural Sarsa and Q-learning, DQN, double and duelling DQN with experience replay and target networks, neural TD(λ) and Sarsa(λ), REINFORCE, n-step actor-critic and proximal policy optimisation (PPO) [11].

Every agent is trained by a single call to .interact(), which runs the experiment and returns learning curves, value maps and trajectory views. A configuration dictionary switches on a visual mode that renders the agent moving through its environment inside a Jupyter notebook as it learns, alongside steps-per-episode and return curves, so that students can watch a policy improve rather than read it only from a curve. The rl.select module repeats runs over seeds and compares algorithms or hyperparameters, which supports the classic experiments of the textbook, such as the error of TD(λ) as a function of step size and λ. The repository includes runnable scripts, notebooks that open directly in Google Colab, and automated tests that train a short run of each stage on every change through continuous integration.

## 3. Illustrative examples

Every algorithm is used in the same way: it is constructed with an environment and hyperparameters, and trained with .interact(). Tabular Sarsa on a gridworld is one line:

```
from rl.tabular import *
from env.grid.tabular import *

sarsa = Sarsa(env=grid(), α=.8, episodes=50, seed=10).interact()
```

The same pattern trains a deep Q-network with experience replay and a target network on a sparse-reward maze; replacing DQN with DDQN, nnSarsa or nnQlearn changes the algorithm and nothing else:

```
from rl.neural import *

dqn = DQN(env=imaze(reward='sparse'), α=1e-4, γ=.98, episodes=30,
          trunk=[(8, 4, 2), (4, 4, 4)], nbuffer=1000, nbatch=64).interact()
```

Linear control on a Gymnasium task uses tile coding, and the visual mode animates the car as it learns to climb the hill:

```
demo = {'plotT': True, 'plotR': True, 'visual': True, 'animate': True}
vsarsa = vSarsa(env=vGymDiscreteS(**MountainCar), α=.1, ε=0,
                episodes=500, seed=1, **demo).interact()
```

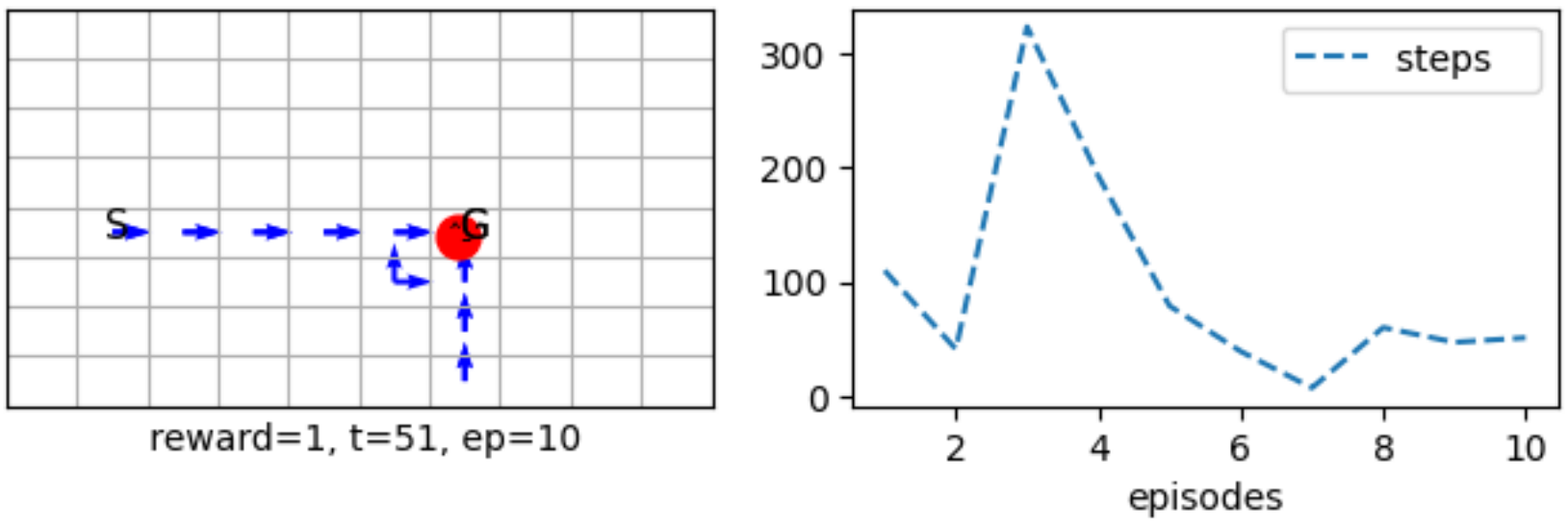

*Fig. 1. Tabular Sarsa on a gridworld, produced directly by .interact(): the learned trajectory (left) and steps per episode (right).*

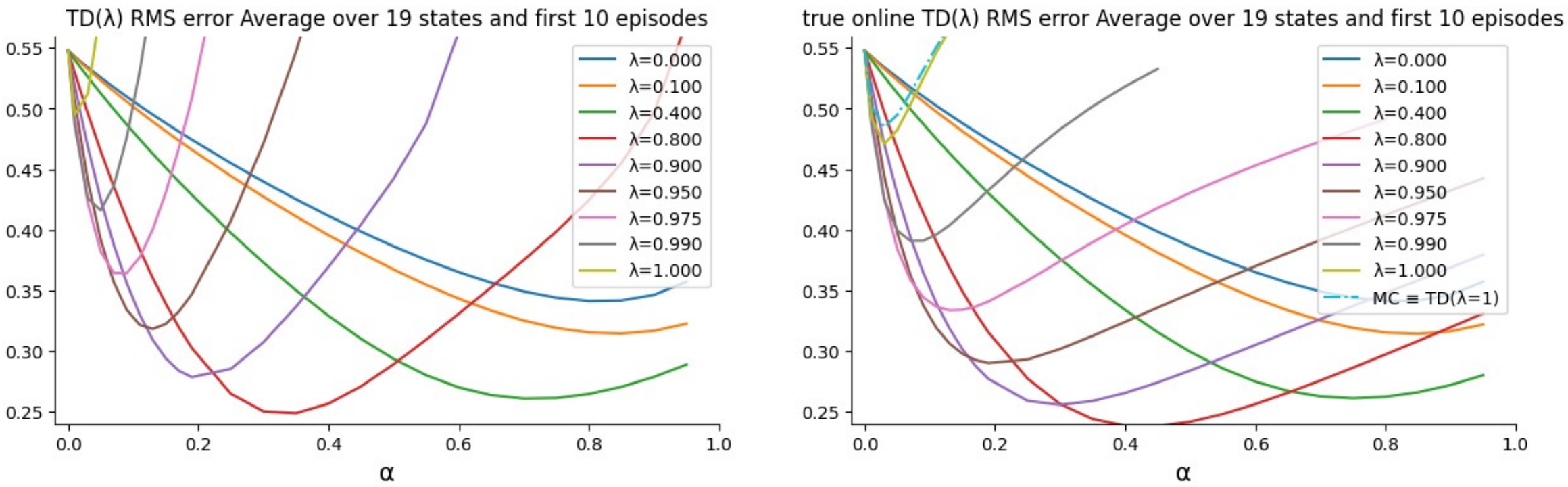

*Fig. 2. Eligibility traces: TD(λ) RMS error against step size α for a range of λ (left) and true online TD(λ) (right), reproduced with the comparison utilities in rl.select.*

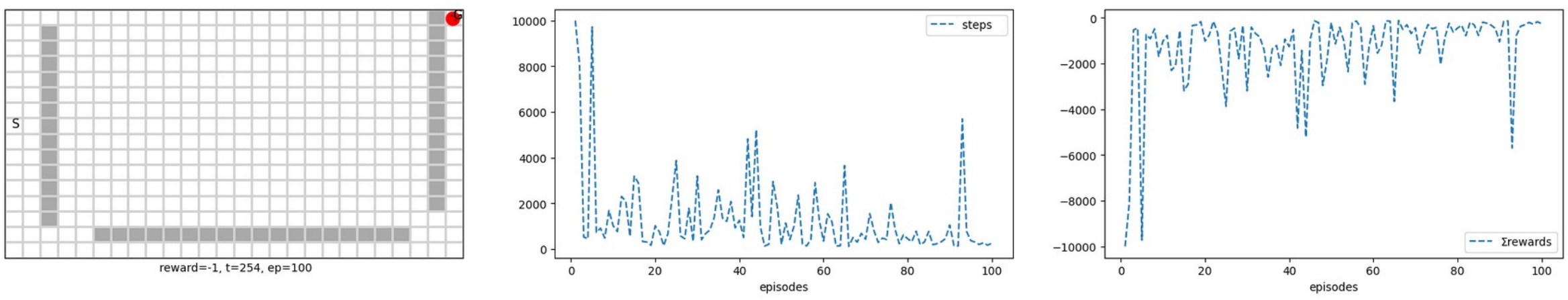

*Fig. 3. Deep value-based control: DQN on a sparse-reward maze, steps and return per episode, trained through the same interface as the tabular agent in Fig. 1.*

Because the interface is shared, a single notebook can place tabular and neural Sarsa side by side on the same maze. Students see directly what function approximation changes, in speed of learning, stability and generalisation, without rewriting the experiment.

## 4. Impact

rl_rob took its present form in the Reinforcement Learning module of the online MSc in Artificial Intelligence at the University of Leeds, a module designed and taught by the author since 2021/22. Its students are mostly part-time working professionals studying online, for whom installation problems and opaque frameworks are a serious barrier. Over five years the library has been used by several hundred postgraduate students.

The module is built around a companion set of Jupyter worksheets that take students from one-step temporal-difference learning to PPO, each with explicit learning outcomes. In each worksheet, students implement or adapt the update rule of an algorithm and then study its behaviour with the library's experiment and plotting tools. The summative assessment mirrors the arc of the module: students take an agent from tabular to linear to non-linear function approximation, measure what each stage gains, meet the slow convergence and divergence they will face in practice, and compare algorithms to choose one that suits their environment. To remove installation barriers, the library also runs in a cloud laboratory and in Google Colab.

Two lessons from this use shaped the design. First, students engage more deeply with an algorithm when the code they write is the update rule itself and the rest of the experiment is handled for them. Second, a single interface across the three stages makes the effect of function approximation concrete, because methods can be compared on the same problem with the same tools. The approach to teaching RL embodied in the library was also the basis of invited tutorials on experience replay and deep RL at the International Joint Conference on Neural Networks, and the library has served as the starting point for doctoral research on RL.

Beyond its home module, rl_rob is released openly so that instructors and self-learners elsewhere can use it alongside the standard textbook [1]. It is suited to postgraduate and advanced undergraduate courses, to self-study, and to instructors who want one code base across a whole module rather than a different framework at each stage.

## 5. Conclusions

rl_rob provides a single, transparent code base for teaching reinforcement learning across its full arc, from tabular methods to deep policy gradients. By writing every update rule explicitly while supplying environments, experiments and visualisation, it lets students study the part of an algorithm that matters for understanding, and lets instructors set tasks that remain meaningful when code generation is freely available. Planned work includes fuller API documentation, further environments, and a structured evaluation of the library's effect on student learning.

## Declaration of competing interest

The author declares that he has no known competing financial interests or personal relationships that could have appeared to influence the work reported in this paper.

## Declaration of generative AI and AI-assisted technologies in the writing process

During the preparation of this work the author used Claude (Anthropic) to assist with drafting and editing the manuscript text. The author reviewed and edited the output as needed and takes full responsibility for the content of the published article.

## Acknowledgements

The author thanks the students of the Reinforcement Learning module of the online MSc in Artificial Intelligence at the University of Leeds, whose questions and feedback shaped the library.